# What Drives Length of Stay After Elective Spine Surgery? Insights from a Decade of Predictive Modeling

**Ha Na Cho, MS[1], Seungmin Jeong, MS[1], Yawen Guo, MS[1], Alexander Lopez, MD, MS[2], Hansen Bow, MD, PhD[2], Kai Zheng, PhD[1]**
**[1]University of California Irvine, Irvine, CA, United States**
**[2]Department of Neurosurgery, University of California, Irvine, Orange, California, United States**

Corresponding author: Ha Na Cho, 6095 Donald Bren Hall, Irvine, CA, 92697-3440, chohn1@uci.edu

## Abstract

**Objective:** Predicting length of stay after elective spine surgery is essential for optimizing patient outcomes and hospital resource use. This systematic review synthesizes computational methods used to predict length of stay in this patient population, highlighting model performance and key predictors.

**Methods:** Following PRISMA guidelines, we systematically searched PubMed, Google Scholar, and ACM Digital Library for studies published between December 1$^{st}$, 2015, and December 1$^{st}$, 2024. Eligible studies applied statistical or machine learning models to predict length of stay for elective spine surgery patients. Three reviewers independently screened studies and extracted data.

**Results:** Out of 1,263 screened studies, 29 studies met inclusion criteria. Length of stay was predicted as a continuous, binary, or percentile-based outcome. Models included logistic regression, random forest, boosting algorithms, and neural networks. Machine learning models consistently outperformed traditional statistical models, with AUCs ranging from 0.94 to 0.99. K-Nearest Neighbors and Naive Bayes achieved top performance in some studies. Common predictors included age, comorbidities (notably hypertension and diabetes), BMI, type and duration of surgery, and number of spinal levels. However, external validation and reporting practices varied widely across studies.

**Discussion:** There is growing interest in artificial intelligence and machine learning in length of stay prediction, but lack of standardization and external validation limits clinical utility. Future studies should prioritize standardized outcome definitions and transparent reporting needed to advance real-world deployment.

**Conclusion:** Machine learning models offer strong potential for length of stay prediction after elective spine surgery, highlighting their potential for improving discharge planning and hospital resource management.

## Introduction

The length of time a patient stays in the hospital after an elective spine operation is important from multiple perspectives. From the patient perspective, longer hospital stays are associated with increased risks of complications, including infection and deep vein thrombosis. Additionally, longer hospital stays are associated with decreased patient satisfaction. For the health care facility, the ability to predict when a postoperative patient will be discharged will help to facilitate capacity management to accept more patients in need of care. Lastly, each additional day of an inpatient hospitalization can cost several thousand dollars, averaging around $2,883 nationally, with variations depending on hospital type and payer, and these expenses are typically borne by either the hospital (in a capitated model) or the patient[1].

Over the past decade, computational methods such as artificial intelligence and machine learning (AI/ML) have advanced exponentially. These methods have been applied to predicting Length of Stay (LOS) after elective spine operations. Inputs to these models may include both structured data (e.g. demographics and clinical metrics) and/or unstructured data (e.g. clinical notes documenting the impressions of healthcare providers). It is unclear currently which ML models are superior with respect to precision and accuracy of the prediction. It is also unclear whether there are added benefits by incorporating unstructured data in the analysis using methods such as natural language processing (NLP).

The integration of data-driven predictive models in the discipline of spine surgery is gaining momentum, with several existing systematic reviews highlighting their potential to optimize patient care. Recent studies have focused on AI/ML models for predicting key outcomes including LOS, readmission rates, and costs. For instance, Lopez et al.[2] found that Bayesian networks outperformed other statistical approaches in length of stay prediction, and Ghanem et al. further identified the importance of proper model evaluation schemes, including sensitivity and specificity, especially for imbalanced datasets commonly found in spine surgery outcomes[3]. Additionally, Lubelski et al.[4] examined 31 studies on LOS prediction but noted lack of external validation, with performance metrics ranging from Area Under the Curve (AUC) values of 0.60 to 0.90, describing the need for further validation before widespread clinical adoption; and Sirocchi et al. emphasized that while AI/ML models hold great potential, their integration into clinical practice remains limited due to challenges in capturing the full clinical context, including nuanced patient histories, which are crucial for accurate predictions[5]. Lastly, Tragaris et al. reviewed 46 studies and reported that AI/ML models demonstrated an average accuracy of 74.9% in preoperative patient selection, with growing interest with such models after 2018[6].

Collectively, these review studies demonstrate the evolving role of AI/ML in spine surgery, particularly in predicting LOS. Notably, only one prior review study specifically addressed prediction models for elective spine surgery[4]. However, Lubelski's review scope 1) examined diverse outcomes such as pain scores, complications, reoperation, and readmission, while not focusing solely on LOS prediction, 2) offered limited comparison between machine learning models and traditional statistical approaches, and 3) did not differentiate which variables and predictors most significantly impacted model performance for the reviewed studies. Lubelski's review addressed nine LOS or discharge outcomes, but none provided a detailed comparative analysis of the performance of predictive models for LOS specifically in elective spine surgery. Our review, in contrast, focuses exclusively on LOS outcomes in elective spine surgery. By analyzing data-driven models, including AI, statistical, and machine learning models, this study addresses critical gaps left by Lubelski et al. and includes 16 additional papers published after 2020. This addition highlights advancements in AI methodologies and the increasing role of data integration, which significantly enhances model performance.

To address these gaps, this review provides robust and comprehensive evaluation of data-driven models for predicting postoperative LOS. A key strength of our study lies in the thorough extraction and analysis of the variables integrated into the predictive models and the specific predictors that impact model performance. Unlike prior review studies[4,7], we systematically collected and categorized the vast array of predictors, such as demographic, clinical, surgical factors, and other unique factors-contributing to more granular insights into model outcomes. Furthermore, our review stands out in its examination of model performance. By comparing the performance scores of various models, including machine learning and statistical approaches, we highlight the best-performing models based on key metrics like AUC, accuracy, and z-score. This exhaustive approach not only identifies the current state of the art but also underscores the limitations and areas for future improvement in LOS prediction models for elective spine surgery. By providing a focused analysis of LOS and discharge prediction models, this study offers actionable insights for clinicians and healthcare systems, guiding the integration of effective data-driven strategies to optimize care delivery and operational efficiency.

## Methods

### Search Protocol

Our systematic review followed the Preferred Reporting Items for Systematic Reviews and Meta-Analysis (PRISMA) guidelines[8]. Studies were included based on the following criteria: (1) Original empirical study predicting LOS in elective spine surgery patients (2) Utilized data-driven techniques for prediction (3) Published in English, peer-reviewed, and with full-text access (4) Studies on non-elective spine surgeries or those not using data-driven prediction techniques were excluded.

### Search Strategy

A comprehensive search was conducted using databases such as PubMed, Google Scholar, and ACM. The search query included combinations of keywords: (“elective spine surgery” OR “planned spine surgery”) AND (“length of stay” OR “postoperative LOS” OR “discharge”) AND (“predictive model” OR “machine-learning” OR “artificial intelligence”). The search strategy was designed to maximize the capture of relevant studies published between December 1st, 2015, and December 1st, 2024.

### Study Selection

Titles and abstracts were screened by two independent reviewers to ensure adherence to the inclusion criteria. Full texts of eligible studies were retrieved and reviewed in detail. Disagreements were resolved through discussion with a third reviewer. The selection process was summarized in a PRISMA flow diagram in Fig. 1.

Figure 1. PRISMA flow diagram illustrating the study filtration process.

**Data Collection**

Two reviewers (HC, SJ) independently screened titles and abstracts, followed by full-text reviews to confirm inclusion based on predefined criteria: author details, year of publication, study location, study design, sample size, model type, predictive features used, LOS outcomes, main findings, and conclusions. Discrepancies were resolved by consensus, with a third reviewer (YG) adjudicating when necessary. Weekly consensus meetings were held at each stage to ensure rigor. Articles were categorized as (1) non-data-driven or (2) data-driven approaches (machine learning and statistical models).

**Codebook Development**

The initial step in our data extraction process was to develop a comprehensive codebook. The codebook served as a detailed framework for systematically categorizing and extracting information from the studies. We identified key domains including study characteristics, modeling approaches, and evaluation metrics. Variables were clearly defined and categorized into domains such as study design, data type, modeling strategy (statistical vs. ML), and evaluation methods. The codebook also captured surgery type, sample size, patient age, variable types, outcome definitions, and performance measures.

**Data Extraction Process**

Using the codebook, we extracted the followings:

*1. Study Characteristics:* Title, first author, journal, and publication year, study design (retrospective or prospective), dataset characteristics, sample size, mean patient age, and surgery type.

*2. Outcome Definition and Classification:* LOS was categorized as continuous, binary, or percentile based. A spine surgeon (BH) confirmed eligibility for elective spine surgery studies. LOS was further classified as “prolonged” or “normal” using study-specific thresholds.

*3. Modeling Approaches:* Modeling approaches were classified as statistical, ML, or NLP. Specific techniques (e.g., regression, random forest, SVM), were documented to track methodological trends.

*4. Variable and Predictor Analysis:* Predictors were categorized into demographics, operational, clinical, and other variables. We documented features used in each model and highlighted key predictors associated with prolonged LOS.

*5. Performance Metrics:* Metrics such as Area Under the Curve (AUC), precision, accuracy, F1 score, and recall were collected to evaluated and compare model performance across studies. The best-performing models were noted.

**Risk of Bias Assessment**

The quality of included studies was assessed using the Cochrane risk of bias tool. Each study was evaluated for potential selection, performance, detection, attrition, and reporting bias[9].

**Results**

Lubelski's review spanned publications from January 1, 2000, to March 1, 2020, identifying 31 articles, of which 9 addressed LOS or discharge outcomes[4]. In contrast, our study included an additional 16 papers published after 2020, reflecting advancements in AI and data-driven models.

**Search Results**

From an initial pool of 1,263 studies identified across databases (890 from PubMed, 199 from Google Scholar, and 174 from ACM), 40 (3.2%) duplicates were removed. The remaining 1,223 (96.8%) studies were screened based on title and abstract, resulting in exclusion of 1,007 (79.7%) studies due to irrelevance. During the full-text assessment, 186 (14.7%) studies were excluded for reasons such as incorrect outcomes, settings, patient populations, and study designs (Fig. 2). Ultimately, 29 (2.3%) studies met the criteria and were included in the data extraction phase.

Figure 2. Number of included studies by publication year range (n=29).

**Study Characteristics**

**Overview**

Our systematic review encompassed a broad spectrum of elective spine surgery procedures, reflecting the diversity of techniques and regional practices in current spinal care. The studies analyzed were predominantly retrospective, comprising 90% of the total, with prospective studies making up the remaining 10%. The data set sizes varied significantly across studies, ranging from fewer than 100 to 450,000 patients, with an average sample size of 29,671. This considerable variance shows the extensive and varied nature of research conducted in this field. Based on the data sources used in the studies, the majority relied on electronic health records (EHR) systems. Specifically, 44.8% utilized Electronic Medical Records (EMR) data from various institutions, including Mount Sinai Medical Center and the University of California, San Francisco. The use of national databases was also prevalent, with 20.7% employing data from the American College of Surgeons National Surgical Quality Improvement Program (NSQIP), while 6.9% utilized the Nationwide Readmissions Database (NRD) and the Centers for Medicare and Medicaid Services (CMS) (Table 1). Other sources included state-level databases like the SPARCS administrative database (3.4%), the Quality Outcomes Database (QOD) for spondylolisthesis (3.4%), and single-center prospective registries, such as those from Bergman Clinics (3.4%).

Most of the studies, accounting for 72.4%, were published between 2020 and 2024, demonstrating a recent surge in research interest and advancements in this area. The remaining 27.6% of papers were published prior to 2020 (Table 1). This trend highlights an increasing focus on developing more predictive models for postoperative outcomes in recent years (Figure 1).

| 1st Author, Year | Data Source | Sample Size | Surgical Procedure | Prediction Model |
|---|---|---|---|---|
| Arora, 2022[10] | EHR* | 3,678 | Any fusion;<br>Decompression<br>(lumber, Cervical, thoracic) | Statistical model;<br>Other |
| Arora, 2023[11] | State-level inpatient database from HCUP* | 8,866 | Multi-level lumbar/thoracolumbar spinal instrumented fusion | Statistical model |
| Basques, 2014[12] | EMR* | 183 | ACDF* | Statistical model |
| Cabrera, 2023[13] | ACS-NSQUP* database | 29,949 | PCDFI* | Machine learning;<br>Statistical model |
| Dandurand, 2022[14] | The multicenter CSORN* | N/A | Thoracolumbar fusion (Discectomy; Laminectomy; PIF*) | Statistical model |
| Fry, 2017[15] | CMS* | 167,395 | Cervical fusion;<br>Non-cervical Fusion | Statistical model;<br>Other |
| Gowd, 2022[16] | NSQIP* database | 42,194 | ACDF | Machine learning |
| Gruskay, 2015[17] | EMR | 103 | PLF* | Statistical model |
| Hagan, 2022[18] | EMR, CPT* | 799 | Lumbar surgery without fusion | Statistical model |
| Hung, 2022[19] | EMR | 240 | Lumbar fusion;<br>ACDF;<br>Deformity surgery | Statistical model |
| Janssen, 2021[20] | EMR | 77 | Lumbar Fusion | Machine learning |
| Kanaan, 2015[21] | EMR | 593 | Lumbar fusion (Arthrodesis);<br>Decompression<br>(Laminotomy, Laminectomy) | Statistical model |
| Karabaca, 2023[22] | NSQIP* database | 3,073 | Tumor resection | Machine learning;<br>Statistical model |
| Karnuta, 2020[23] | SPARCS administrative database | 38,070 | Lumbar fusion | Machine learning |
| Li, 2024[24] | Degenerative Spine Data | 162 | Posterior fusion | Machine learning;<br>Statistical model |
| Lubelski, 2020[25] | EMR | 257 | Decompression<br>(Lumbar, Thoracic); | Statistical model |
| | | | Lumbar fusion; ACDF; PCF* | |
| Martini, 2020[26] | Single academic health system consisting of multiple hospitals and surgeon practices | 11,150 | Any surgery for degenerative conditions (Cervical, Thoracolumbar) | Machine learning |
| Meng, 2017[27] | EMR; UC San Francisco | 102 | Lumbar or thoracolumbar spine surgery (Posterior) | Statistical model |
| Moskven, 2024[28] | CSORN | 1,228 | ACDF; PCF | Statistical model |
| Passias, 2018[29] | NSQIP | 60,964 | Any region; Decompression (Laminectomy, Discectomy, Laminoplasty); Fusion (Arthrodesis) | Statistical model |
| Sachdeva, 2020[30] | Academic tertiary care referral center | 81 | LLIF*; TLIF* | Statistical model |
| Senker, 2021[31] | Clinical data | 187 | Lumbar fusion (Minimally invasive) | Statistical model |
| Shahrestani, Jun 2023[32] | QOD* spondylolisthesis data set | 544 | Decompression alone; Decompression plus fusion | Machine learning |
| Shahrestani, Dec 2023[33] | NRD* | 453,717 | Cervical fusion | Statistical model |
| Siccoli, 2019 [34] | Bergman Clinics' prospective registry | 635 | Decompression for LSS* | Machine learning |
| Valliani, 2022[35] | EMR | 4,400 | Cervical Fusion | Machine learning |
| Wang, 2024[36] | Geriatric Lumbar Disease Database | 1,020 | TLIF | Machine learning; Statistical model |
| Yashin, 2023[37] | EMR | 540 | Decompression for LLS; TLIF | Machine learning; Statistical model |
| Yuk, 2017[38] | EMR; Mount Sinai Medical Center | 587 | Cervical surgery (Anterior, Posterior) | Statistical model |

Table 1. Overview of included study characteristics.

*ACDF: Anterior cervical discectomy and fusion

*ACS-NSQIP: American College of Surgeons National Quality Improvement Program

*CMS: Centers for Medicare and Medicaid Service

*CSORN: Canadian Spine Outcomes and Research Network

*EHR: Electronic Health Record

*EMR: Electronic Medical Record

*HCUP: Healthcare Cost and Utilization Project

*LLIF: Lateral lumbar interbody fusion

*LSS: Lumbar spinal stenosis

*NSQIP: National Surgical Quality Improvement Program

*NRD: Nationwide Readmissions Database

*PCDFI: Posterior cervical decompression and fusion with instrumentation

*PCF: Posterior cervical fusion

*PIF: Posterior instrumented fusion

*PLF: Posterior lumbar fusion

*QOD: Quality Outcomes Database

*TLIF: Transforaminal lumbar interbody fusion

Patient demographics revealed a mean age of 60 years, indicating a predominantly older patient population undergoing these procedures. The types of elective spine surgeries examined primarily focused on lumbar and cervical spine surgeries. Lumbar spine procedures, such as lumbar spinal fusion and transforaminal lumbar interbody fusion, were the most prevalent (34.5%). Cervical procedures, including anterior cervical discectomy and fusion (ACDF) and posterior cervical fusion, were featured in 27.6% of the studies. Multi-level fusions, like thoracolumbar and multi-level lumbar instrumented fusions, accounted for 13.8%, while minimally invasive lumbar fusion techniques and decompression surgeries each represented 10.3%. Less common were specialized surgeries, such as spinal tumor resections and deformity corrections, at 6.9%. The focus on lumbar and cervical regions and diverse surgical techniques, including ACDF, posterior cervical laminectomy and fusion, lumbar laminectomy, lumbar fusion, deformity correction, and tumor surgery, reflects the emphasis on addressing elective spine conditions and optimizing recovery times.

### Outcome

The specific outcome of LOS was variably defined across the studies, highlighting the complexity of postoperative care assessment. Definitions ranged from counts of days from surgery to discharge, to more nuanced measures involving percentiles and specific postoperative complications. For some studies, LOS was simply recorded as the number of days from surgery until discharge to home or rehabilitation, often described in whole days. Others specified prolonged LOS as being greater than a certain threshold, with further distinctions at more than the 75th percentile, or defining extended LOS as beyond the top quartile (e.g., greater than 4 days). Several studies adopted a percentile-based approach to categorize LOS, identifying “extended” or “prolonged” stays as exceeding the 75th or even the 90th percentile, reflecting significant deviations from typical recovery patterns. The studies in our review utilized three types of outcome variables: continuous, binary, and categories.

### Models Used

Of the 29 included papers, 17 (58.6%) studies utilized statistical models, such as logistic regression and linear regression, as their primary approach. Logistic regression model was most used, with 68.97% (20/29) studies, often serving a baseline due to its simplicity and interpretability. On the other hand, 7 (24.1%) studies employed ML methods, including Random forests (20.69%), gradient boosting techniques like XGBoost and LightGBM (17.24%), support vector machines, and k-nearest neighbors (KNN) (6.90%) (Table 2). While statistical methods dominated the field, ML approaches were increasingly utilized for their ability to handle complex, non-linear relationships and improve predictive performance. Additionally, five studies (17.24%) compared statistical and ML methods, showing hybrid approaches that integrated logistic regression with models like random forest or boosting tree models. As summarized in Fig. 3, logistic regression was the best-performing model in nearly half of the studies, followed by linear regression, random forest and ensemble-based models.

| 1st Author, Year | Model name | Performance: AUC | Best Performed Model |
|---|---|---|---|
| Arora, 2022[10] | Regression, Classification Trees, Least Absolute Shrinkage and Selection Operator | AUC=0.79, Sensitivity=0.80, Specificity=0.64<br>LASSO: 0.79 (Discharge disposition)<br>Decision Tree Learning: 0.69 (Discharge disposition) | Logistic regression |
| Arora, 2023[11] | Logistic regression | AUC=0.77 | Logistic regression |
| Basques, 2014[12] | Multivariate logistic regression, bivariate logistic regression | Average LOS=2.0±2.5 days | Multivariate logistic regression |
| Cabrera, 2023[13] | Random Forest, t-tests, ANOVA | Accuracy=0.78, recall=43.33 | Random forest |
| Dandurand, 2022[14] | Logistic regression models | AUC=0.84 | Logistic regression |
| Fry, 2017[15] | Logistic regression models | A total of 54 hospitals (6.2%) had z-scores that were 2.0 better than predicted with a median risk adjusted AO rate of 9.2%, and 75 hospitals (8.6%) were 2.0 z-scores poorer than predicted with a median risk-adjusted AO rate of 23.2%. | Logistic regression |
| Gowd, 2022[16] | Logistic regression, gradient boosting trees, random forest, decision tree | AUC=0.73 | Logistic regression |
| Gruskay, 2015[17] | Multivariate linear stepwise regression, Pearson bivariate cross-correlation | Average LOS=3.68 | Multivariate linear regression |
| Hagan, 2022[18] | Multivariable logistic regression, Ordinal logistic regression | AUC=0.87 | Multivariable logistic regression |
| Hung, 2022[19] | Wilcoxon rank-sum test, Fisher's exact test, multivariable logistic regression | AUC=0.69 | Multivariable logistic regression |
| Janssen, 2021[20] | Random forest | Z score=5.1 | Random forest |
| Kanaan, 2015[21] | Mahalanobis d-squared distance, Regression model, Q-Q plots, univariate normality distribution | Average LOS=4.01 | Logistic regression |
| Karabaca, 2023[22] | XGBoost, LightGBM, CatBoost, random forest, Welch t-test, Mann Whitney U test, Pearson, chi-squared test, Shapiro test, | AUC=0.74<br>LGB=0.75<br>CB=0.73<br>RF=0.76 | Random forest |
| | Levene test | Mean=0.75 | |
| Karnuta, 2020[23] | Naive Bayes with adaptive boosting | AUC=0.94 | Naive Bayes |
| Li, 2024[24] | Logistic Regression, Decision Tree, Elastic Networks (Enet), K Nearest Neighbors (KNN), Light Gradient boosting machine (Lightgbm), Random Forest, eXtreme Gradient Boosting (XGBoost), support vector machines (SVM), Multilayer perceptron (MLP), Pearson Chi-square tes, t test, Wilcoxon rank test | AUC=0.81 | K-nearest neighbors |
| Lubelski, 2020[25] | Logistic regression | AUC=0.89 | Multivariable logistic regression |
| Martini, 2020[26] | Gradient Boosting Trees, Shapley additive explanation | Mean C-statistic=0.87 | Gradient Boosting Trees |
| Meng, 2017[27] | Univariate Analysis, Multiple Linear Regression Models, Student's t-test | P-value=0.31(+19.0%) | Linear regression |
| Moskven, 2024[28] | Multivariable logistic regression | AUC=0.71<br>PCF: 0.75 (95% CI 0.71-0.80, p<0.001) | Logistic regression |
| Passias, 2018[29] | ANOVA t-test | N/A | ANOVA t-test |
| Sachdeva, 2020[30] | Linear regression model, ANOVA | Mean LOS: (fuse level) 1= 2.47; 2= 3.35; 3= 2.68; 4= 3.91 | Logistic regression |
| Senker, 2021[31] | Multivariate logistic regression, Pearson correlation, Spearman rank correlation | N/A | Pearson correlation |
| Shahrestani, Jun 2023[32] | Mann-Whitney U-tests for variable selection, k-nearest neighbors (KNN) models | AUC=0.99, sensitivity=1, specificity=0.85 | K-nearest neighbors |
| Shahrestani, Dec 2023[33] | Elixhauser Comorbidity Index | Average LOS=2.7±4.4 days | Elixhauser Comorbidity Index |
| Siccoli, 2019 [34] | XGBoost, random forest, Bayesian linear | AUC=0.58, accuracy=0.77 | Gradient Boosting Trees |
| Valliani, 2022[35] | Logistic regression, Gradient Boost, Shapley additive explanation | AUC=0.87 | Gradient Boosted trees |
| Wang, 2024[36] | Logistic regression model, Single classification and regression tree, Random Forest model | AUC=0.73 | Logistic regression |
| Yashin, | Multivariate logistic regression, | AUC=0.81 | Multivariate |
| 2023[37] | LASSO, Q-Q plots, Student t-test, Mann-Whitney U-test, chi-square test, fisher exact test | | logistic regression |
| Yuk, 2017[38] | GraphPad Prism 5 and SPSS, Unpaired t-tests, Mann-Whitney test, chi-squared tests, Fisher exact tests, Multiple linear regression | N/A | Multivariable linear regression |

Table 2. Overview of models and best-performing approaches.

Figure 3. Distribution of best performing predictive models.

**Model Performance and Evaluation Metrics**

Performance was primarily evaluated using AUC, with values ranging from 0.70 to 0.99 scores across the studies. Additional evaluation metrics included accuracy, precision, F1 score, and recall, with ML models generally reporting higher AUC values compared to statistical models, particularly in studies applying ensemble methods. Notably, KNN and Naive Bayes with adaptive boosting demonstrated the optimal performance ranging from 0.94 and 0.99 AUC, followed by random forest and Gradient Boosting models providing superior performance in multiple studies, frequently exceeding 0.80 AUC. Logistic regression models showed respectable AUC values, often ranging between 0.70 and 0.80. Studies using more complex models, such as support vector machines, reported competitive results, though their performance depended heavily on the quality, class imbalance, and diversity of the input data.

**Variables and Predictors**

The models used in these studies incorporated a wide range of variables, including demographic (e.g., age, sex, race/ethnicity, education levels, living status, marital status, insurance status, Zip Code), clinical (e.g., BMI, weight, height, ASA score, comorbidities, hematologic conditions, symptom parameters, PROMs, pain medication use, physical/neurological conditions), operational (e.g., diagnosis, admission types, admission date, preoperative days, surgical specialty, surgical procedure, surgical approaches, surgical regions, surgical levels, incision length, surgeon experience, surgery history, operation duration, operation date, adjunctive medication use, transfusion, fluid/drain management, ventilator use, complications, discharge positions, mobility/ambulation, protocol to reduce LOS, readmission, reoperation, postoperative therapy), and other contextual factors (e.g., substance use, impatient cost, ADI, healthcare institution type) (See Supplementary Appendix 1). Age emerged as the most significant demographic predictor, included in 79.3% (23/29) of the studies, with older patients generally experiencing longer LOS due to slower recovery rates. Clinical variables, particularly comorbidities like diabetes, hypertension, and cardiovascular conditions, were strong predictors, included in 79.3% (23/29) of studies and contributing to prolonged hospital stays in 65.5% (19/29) (See Supplementary Appendix 2). Surgical factors such as duration and estimated blood loss (EBL) were critical, with longer operative times and higher EBL linked to extended LOS in 72.4% (21/29) and 55.2% (16/29) of studies, respectively. Other variables, such as smoking status (27.6%) influenced recovery time and discharge readiness, though less consistently. Additionally, 62.1% (18/29) of studies found that minimally invasive procedures were associated with shorter LOS compared to open surgeries.

**Discussion**

This systematic review included 29 articles focused on predicting LOS for patients undergoing elective spine surgery. The included studies applied various data-driven modeling approaches, ranging from traditional statistical models to ML techniques. Our findings indicate that statistical models, such as logistic regression, were more commonly employed compared to other ML models. The studies reviewed utilized two main approaches to modeling: using a single type of model or comparing multiple models. Studies that compared various models generally found ML models to outperform statistical approaches, especially in handling high-dimensional data and complex interactions among variables. Methods such as KNN and Naive Bayes with adaptive boosting demonstrated top performances in these studies.

Our analysis of predictors for LOS in elective spine surgery reveals the complexity of factors that affect patient recovery times. Understanding these predictors is crucial for improving discharge planning, optimizing resource allocation, and enhancing patient care. Age consistently emerged as a key predictor, with older patients generally facing longer hospital stays due to reduced physiological resilience and greater risks of complications, caused by comorbidities such as cardiovascular conditions and diabetes. These comorbidities were frequently linked to prolonged LOS, underscoring the need for predictive models to account for patients' overall health status. The inclusion of such clinical factors allows for a more tailored approach to postoperative care, helping to identify patients who may require extended monitoring and support.

Surgical factors, such as duration and intraoperative blood loss, also significantly influenced LOS. Longer operative times and higher blood loss often signal more complex procedures, necessitating additional recovery time and resources. This insight emphasizes the importance of integrating these variables into models to improve predictions and optimize resource allocation. While factors like smoking status and socioeconomic status were not as consistently studied, their impact suggests that considering lifestyle and social determinants could further enhance the precision of discharge planning[39]. Ultimately, these findings advocate for a comprehensive approach in predictive modeling that incorporates diverse clinical, demographic, and operational data, enabling more effective management of LOS and improving outcomes in elective spine surgery.

In terms of model performance, The AUC was the most frequently reported evaluated metric, reflecting the models' ability to distinguish between outcomes effectively. AUC values across the reviewed studies indicate a wide spectrum of predictive accuracy. Sensitivity and specificity were often reported alongside AUC to provide a more comprehensive assessment of model performance. Moreover, in our review, most studies reported AUC values within the range of 0.70 to 0.80, reflecting moderate to strong predictive performance in line with the typical complexity of LOS study using surgery datasets. Notably, AUC values exceeding 0.90 were rare, where one study reported an exceptionally high value of 0.99. While this result is impressive, it is worth exploring further, as such near-perfect performance is uncommon in real-world applications. It may suggest unique factors at play, such as unusually homogeneous dataset used, the presence of highly predictable features, or potential methodological artifacts like data leakage issues[40].

For example, datasets with low variability or class imbalance might also contribute to inflated AUC scores, if not adequately preprocessed.

Recent state-of-the-art studies have shown the potential of hybrid deep learning models in LOS prediction across various medical fields, which is also valuable for the future direction of prediction model study in spine surgery research. For instance, a study by Neshat et al.[41] introduced a robust model combining Convolutional Neural Networks (CNNs), Gated Recurrent Units (GRU), and Dense Neural Networks, which outperformed other models like LSTM and BiLSTM, achieving an average accuracy of 89% across cross-validation tests. Similarly, a study on Medic-BERT (M-BERT) utilized a modified version of the BERT model to predict LOS using patient event sequences from EHR[42]. The model effectively captured long-term dependencies between medical events, incorporating patient demographics and clinical details as part of the input. This approach demonstrated improved performance by integrating disparate data types, such as diagnostic and therapeutic events, and applying self-attention mechanisms, which significantly enhanced the prediction accuracy of LOS. Another study, GRAPHCARE: Enhancing Healthcare Predictions with Personalized Knowledge Graphs[43], introduced a bi-attention augmented Graph Neural Network (GNN) to predict LOS. This model utilized personalized knowledge graphs (KGs) generated for each patient, incorporating both visit-level and node-level attention to effectively capture complex relationships between patient features. By leveraging both clinical events and medical concepts, GRAPHCARE demonstrated improved accuracy in predicting LOS, highlighting the effectiveness of attention mechanisms in personalized healthcare modeling. These findings align with our review's observation that models leveraging both temporal and convolutional features, as well as advanced architectures like LLMs and GNNs, tend to capture nonlinear patient recovery trends more effectively. It showcases the potential implementation of hybrid modeling for LOS prediction of spine surgery,

Furthermore, future studies should implement advanced AI/ML technology, such as Large-Language Model (LLM) technology, to include unstructured clinical data such as clinical notes, operative reports, discharge summaries, and radiology reports, into the modeling process for predicting elective spine surgery discharge. These datasets can provide valuable insights by capturing detailed patient-specific information, clinical decision-making, and contextual factors often absent in structured data. The inclusion of such textual data could significantly enhance the modeling performance by utilizing advanced natural language processing techniques and large language models. As these technologies continue to evolve, they hold the potential to extract deeper patterns and relationships within the dataset, thereby improving the accuracy and reliability of length of stay predictions.

The reviewed studies exhibited several limitations that must be considered when interpreting their findings. First, many studies lacked robust model external validation, which is a critical step in assessing the model's generalizability to new and independent datasets. Without external validation, the reliability of these predictable models across diverse clinical settings remains uncertain, potentially limiting their practical utility in real-world applications. Small sample sizes in several studies also affected the stability and reliability of the predictive models, as models built on limited data may overfit and fail to perform well on new patients. Moreover, inconsistencies in definitions of the LOS outcome and variability in predictor further complicated the interpretation of findings, making direct comparison of model performance across studies difficult. These limitations emphasize the need for future research to address these gaps by ensuring robust validation processes, and standardized outcome definitions.

**Conclusions**

This review highlights the changing field of predictive modeling for LOS in elective spine surgery, with a growing shift toward adopting advanced ML models. Addressing gaps such as the lack of external validation and the limited use of data types is essential for building models into effective tools for enhancing patient care and optimizing resource management. These efforts would ultimately contribute to better patient outcomes and more efficient resource allocation in spine surgery.

**Author Contributions**

Ha Na Cho led the overall study design, developed the review methodology and data analysis plan, and drafted the manuscript. Seungmin Jeong created tables, study screening, data extraction, and data organization. Yawen Guo conducted study screening and data extraction. Alexander Lopez and Hansen Bow provided clinical knowledge and domain-specific insights. Hansen Bow and Kai Zheng contributed to the conceptualization of the study, refined the review framework, and provided supervision and guidance throughout the study process. All authors critically reviewed the manuscript and approved the final version.

## Supplementary Material

Supplementary material is available at Journal of the American Medical Informatics Association online.

## Funding

This work did not receive any funding.

## Competing Interests

There are no conflicts of interest among the authors.

## Ethics Approval

This study was approved by the Institutional Review Board of the University of California, Irvine (IRB# 4537). All data used were de-identified prior to analysis and complied with relevant ethical guidelines and regulations.